# Sentiment Analysis on Bangla and Romanized Bangla Text (BRBT) using Deep Recurrent models

Asif Hassan, Mohammad Rashedul Amin, Abul Kalam Al Azad[a], Nabeel Mohammed[b]
[a]abul.azad@ulab.edu.bd
[b]nabeel.mohammed@ulab.edu.bd
Dept. of Computer Science and Engineering
University of Liberal Arts Bangladesh (ULAB)
Dhaka, Bangladesh

*Abstract*— Sentiment Analysis (SA) is an opinion mining study analyzing people's opinions, sentiments, evaluations and appraisals towards societal entities such as products, services, individuals, organizations, events, etc. Of late, most of the research works on SA in natural language processing (NLP) are focused on English language. However, it is noted that Bangla does not have a proper dataset that is both large and standard. As a result, recent research works with Bangla in SA have fallen short to produce results that can be both comparable to works done by others in other languages and reusable for further prospective research. In this work, a substantial textual dataset of both Bangla and Romanized Bangla texts have been provided which is first of this kind and post-processed, multiple validated, and ready for SA implementation and experiments. Further, this dataset have been tested in Deep Recurrent model, specifically, Long Short Term Memory (LSTM), using two types of loss functions – binary cross-entropy and categorical cross-entropy, and also some experimental pre-training were conducted by using data from one validation to pre-train the other and vice versa. Lastly, the results along with analysis are presented in this research.

## I. INTRODUCTION

Bangla is spoken as the first language by almost 200 million people worldwide, 160 million of whom are Bangladeshi [1]. Bangladeshi people are found to get increasingly involved in online activities such as - getting connected to friends and families through social media, expressing their opinions and thoughts on popular micro-blogging and social networking sites, sharing opinions and thoughts by means of comments on online news portals, doing online shopping through online marketplaces and other such applications. However, it is becoming increasingly harder for such businesses to monitor and analyze market trends, especially when it is done by analyzing the reaction of the customers on their products or services, due to less or no human-to-human interaction in such businesses. Moreover, the task of going through comments and reviews from each individual customers and figuring out the sentiments within is tedious and in some cases simply intractable, especially considering that - usually very high volume of data is generated very quickly in this day and age of digital connectivity. Therefore, application of automated Sentiment analysis (SA) [2] can play a vital role here for enhancing efficiency and productivity.

SA is widely employed as a machine learning application in many areas, and is known by by many other terms e.g. opinion extraction, sentiment mining, opinion mining, subjectivity analysis, emotion analysis, review mining, etc. Most of the research works found on SA are based on the English language, while Bangla SA is still at a formative stage. An interesting work by Das and Bandyopadhyay [2] on subjectivity detection included Bangla but it is not self-sufficient, as English is also needed. However, none of the works truly considered Bangladesh's perspective. We need to consider not just standardized Bangla, but Banglish (Bangla words mixed with English words) and Romanized Bangla. These three major types can again be loosely categorized in - good, standard, bad, wrong, totally wrong, particular to specific location (almost arcane), etc., depending on the level of clarity, grammatical correctness, meaningfulness, personal idiosyncrasies, impact of localization etc. Moreover, for the Romanized Bangla the added complexity is due to the variation in transliteration between people who know English well and those who do not [3]. The reason, that no clear standard is followed when 160 million Bangladeshi people write in any of the mentioned types, makes it all the more complicated and challenging to work with.

In the recent past, Deep Learning methods, specifically recurrent model-based deep learning models have enjoyed a lot of success in Natural Language Processing (NLP), compared to more traditional machine learning methods [4]. While there are other approaches to SA, in this research we will concentrate exclusively on deep learning based techniques. Our key contributions cover –

- A Data set of 10,000 Bangla and Romanized Bangla text samples, where each sample was annotated by two adult Bangla speakers
- Pre-processing the data in a way so that it is readily usable by researchers.
- Application of deep recurrent models on the Bangla and Romanized Bangla text corpus.
- Pre-train dataset of one label for another (and vice versa) to see if it gives better results.

The paper is organized as follows. In section 2, we discussed the background of our work and the works of others in the same field that inspired and helped us in a way. In section 3, we discussed in details about the dataset that we used for our experiments. Section 4 discusses the methodology and also includes the experimental setup for the deep recurrent models. Section 5 has all the discussion about various results found from our experimentation, and lastly the article concludes with section 6.

## II. BACKGROUND

### A. Sentiment Analysis

A key point of our work is Sentiment Analysis, on Bangla (and Romanized Bangla) language. Although the term "Sentiment Analysis" may have appeared for the first time in Nasukawa and Yi [5], research works on sentiment appeared as early as in 2000 [6-8]. With advent of social media on internet e.g. Facebook, Twitter, forum discussions, reviews, and its rapid growth, we were introduced to a huge amount of digital data (mostly opinionated texts e.g. statuses, comments, arguments etc.) like never before, and to deal with this huge data the SA field enjoyed a similar growth. Since early 2000, sentiment analysis has become one of the most active research areas in NLP.

However, most of the works are highly concentrated on English language, favored by the presence of standard data sets. Standard datasets allow researchers to do their own experiments and compare their contributions with those of others. For the English language, an example of such a standard SA dataset is the IMDB Movie Review Data set, which contains 50,000 annotated (positive or negative movie review) movie reviews made by the viewers. This dataset was originally created by Maas, Daly [9] and since then has been used by a multitude of different studies.

A detailed survey paper [10] presented an overview on the recent updates in SA algorithms and applications, categorizing and summarizing total 54 articles that had been published till 2014. Godbole, Srinivasaiah [11] collected opinions from newspaper and blogs, and assigned scores indicating positive or negative opinion to each distinct entity in the text corpus to do SA. In [12], they proposed and investigated a paradigm to mine the sentiment from a popular real-time micro-blogging service like Twitter, and they fashioned a hybrid approach of using both corpus-based and dictionary-based methods in determining the semantic orientation of the tweets.

### B. Sentiment Analysis for Bangla

It is quite unfortunate that there is no standard collection of data, such as - the IMDB dataset, Twitter corpus etc. for Bangla texts. One effort for standardization came from an automatic translation of positive and negative words of SentiWordNet [13]. However, no corpus was created from this work, thereby limiting its usage to word level determination of sentiment, rather than the more complex natural language processing methods. Additionally, such simplified techniques do not consider the variety of ways in which people usually write, e.g. spelling mistakes, using colloquial terms etc.

A small dataset of Bangla Tweets were collected along with Hindi and Tamil by Patra, Das [14], where the authors reported on the outcome of a shared Sentiment Analysis task of Indian languages. They used 999 Bangla tweets for training and 499 for testing. They did some post processing such as pruning of emoticons from the tweets and removal of duplicated posts. This data was annotated manually by native speakers. However, in terms of usability the dataset's small size is a limiting factor for modern deep learning techniques.

Another similar collection was done in [15], where 1400 Bangla Tweets were collected automatically. However, their dataset is not publicly available, and the size of the dataset is rather small.

A slightly larger corpus was collected, automatically annotated and manually verified by Das and Bandyopadhyay [2], as their collection was almost 2500 Bangla text samples from news items and blog posts. The uniqueness of their collection over the ones collected by others [14, 15] was the average size of 288 words of their samples, which is quite a bit larger than the 144 character Tweet limit.

With most of the other works proceeded in the similar way, the two biggest issues with the current state of affairs in Bangla SA research are - first and foremost, the absence of a standard and big enough dataset to compare against, which makes comparison of research work extremely difficult, and secondly, none of the Bangla SA research takes into account the very prominent practical aspect of the use of *Romanized Bangla* [3].

### C. Deep learning

AI (Artificial Intelligence) has been traditionally done in two ways – i) Knowledge based, and ii) Representation learning based. Knowledge base approach to AI uses logical inference rules to reason about statements input by users. Cyc was one of the most famous of such projects [16]. The failure of knowledge based approach was the driving force into finding a way to give AI the ability to gather its own knowledge by extracting patterns or learning from the data – popularly known as Machine Learning. This new algorithm was based on representation of data or feature. That is, the system is given a number of features about the task in hand on which it will give a decision. Clearly if any of the features were wrong, it would mean wrong representation of the data and the system would not perform well. To rectify this situation representation learning based [17] algorithm was used. This algorithm gave better results than the manually tailored representation of data, and allowed systems to adapt to new tasks with ease. However, using this algorithm it was required that high level abstract features from the raw data were extracted without any error caused by misinterpretation due to the factors of variation, as there can be such factors (e.g. an accent in speakers speech) which would cause false representation in absence of highly sophisticated (human like) understanding. However, deep learning performed better with this issue, as it provides with complex representations expressed in terms of a number of other simpler representations.

*D. Recurrent Neural Network*

Recurrent Neural Network or RNN in short, has been widely used in speech recognition, handwriting recognition, natural language processing and others. Moreover, RNN is the precursor to LSTM. While traditional neural networks failed to create a persistent model that would somewhat mimic the way our memory cells work for learning and remembering information, RNN – a class of ANN, has an interesting model design with a loop used as a feed-back connection which makes the information persistent [18, 19]. The loop enables the flow of information from one step to the next. It is like there are multiple copies of same network, where a successor gets information from all the predecessors, connected in architecture that excels at processing sequential data.

*E. Long Short Term Memory (LSTM)*

While RNN's success was critical in speech and pattern recognition due to its ability to remember temporal dependencies, it was not without problems. RNNs were able to connect previous information to current task, only when the gap between the information was small. As the gap widened, RNNs started to perform poorly. Also, the depth and complexity of layers are increase, the vanishing gradient problem causes difficulty in training. Long Short Term Memory (LSTM) is an extension of simple RNNs, which reduce the vanishing gradient problem and can remember dependencies over larger gaps [20]. In 1997, Hochreiter and Schmidhuber introduced LSTM, where a memory cell had linear dependence of its present activity and its past activity. Input and output gates were introduced to efficiently modulate input and output. However, the introduction of forget gates were crucial to effective modulation of the information flow between present and past activities. [21, 22].

$$i_t = \sigma(W_{xi}x_t + W_{hi}h_{t-1} + W_{ci}c_{t-1}) \quad (1)$$

$$f_t = \sigma(W_{xf}x_t + W_{hf}h_{t-1} + W_{cf}c_{t-1}) \quad (2)$$

$$c_t = f_t \odot c_{t-1} + i_t \odot tanh(W_{xc}x_t + W_{hc}h_{t-1}) \quad (3)$$

$$o_t = \sigma(W_{xo}x_t + W_{ho}h_{t-1} + W_{co}c_{t-1}) \quad (4)$$

$$h_t = o_t \odot tanh(c_t) \quad (5)$$

Equations 1-5 capture the LSTM model where, σ is the logistic sigmoid function. *i, f, o* and *c* are the input gate, forget gate, output gate, and memory cell activation vectors, respectively. The process in LSTM includes three gating functions. Each memory cell $c_t$ has its net input modulated by the activity of an input gate, and has its output modulated by the activity of an output gate. These input and output gates provide a context-sensitive way to update the contents of a memory cell. The forget gate modulates amount of activation of memory cell kept from the previous time step, providing a method to quickly erase the contents of memory cells.

III. DATASET DETAILS

Our dataset is called the BRBT dataset where BRBT stands for Bangla and Romanized Bangla Texts. This Bangla Sentiment Analysis (SA) dataset consists of total 9337 post samples. The dataset is unique because not only this is larger compared to others, but it also encompasses the till-now-ignored Romanized Bangla. Romanized Bangla is the Bangla written in English alphabets. Inclusion of Romanized Bangla in the dataset is paramount, because the ease of writing Bangla using any standard QWERTY keyboard (without a Bangla keyboard e.g. Bijoy® keyboard) and the simplicity of using English as base language for the posts, have popularized Romanized Bangla not just in personal messages and micro-blogs but also in Govt. sanctioned mass messages/announcements. The dataset is currently kept private for safe keeping and further improvement. However, it may be made available by personally contacting the owner/authors, and signing a consent form.

*A. Data Statistic*

Bangla texts holds 72% of whole textual data in the dataset while Romanized Bangla texts is the remaining 28%. There are –

- Total number of entries: 9337
- Bangla entries: 6698
- Romanized Bangla entries: 2639

*B. Data Sources*

Data were collected from various micro-blog sites, such as, Facebook, Twitter, YouTube etc, and some online news portal, product review panels etc. Following is the statistic of data sources -

- From Facebook: 4621
- From Twitter: 2610
- From YouTube: 801
- From online news portals: 1255
- From product review pages: 50

*C. Post collection data processing*

- *Removal of emoticons:-* emoticon, hash-tags were removed to give annotators an unbiased-text-only content to make a decision based on three criteria - positive, negative and ambiguous.

- *Removal of proper nouns:-* Proper nouns were replaced with tags to provide ambiguity. All text samples were collected from publicly available sources and did not reflect the opinion of the authors.

- *Manual validation (by native speakers):-* Collected data samples are manually annotated into one of three categories: positive (1), negative (0) and ambiguous (A). Each text sample was independently manually annotated by two different native Bangla speaking individuals for total two validations. Each annotator validated the data without knowing decisions made by other. This ensures that the validations are unbiased and personal.

TABLE I. DATASET VALIDATION SAMPLES

| Text Sample | Translation | 1st Annotator | 2nd Annotator |
|---|---|---|---|
| অনেক ভালো হয়েছে গান! | Very nice song! | Positive | Positive |
| মর্মান্তিক সড়ক দুর্ঘটনায় ৩ জন নিহত। | 3 dead in a tragic road accident. | Negative | Negative |
| Chotobelar modhur din gulo khub miss kori | Really miss the sweet childhood days | Positive | Negative |
| Sympony er set gula kemon? | How are Symphony mobile sets? | Positive | Ambiguous |
| আলো আলো তুমি কখনো আমার হবেনা | Light, light, you'll never be mine | Ambiguous | Negative |

*D. Double Validation Analysis*

Table 2 gives shows the confusion matrix between the labels given by the two annotators. We can see that the annotators agreed on 75% of the texts samples, giving us a base-line of human level agreement for this data set. Not surprisingly, the greatest amount of disagreements arise on text samples which at least one of the annotators labelled as ambiguous.

TABLE II: CONFUSION MATRIX OF MANUAL ANNOTATIONS

| First Validation | Second Validation | | |
|---|---|---|---|
| | Positive | Negative | Ambiguous |
| Positive | 2817 | 538 | 392 |
| Negative | 178 | 3864 | 404 |
| Ambiguous | 27 | 95 | 1022 |

## IV. DATASET SETUP

The data was manually picked from various online micro-blog sites, product review panels, news portals etc. For tweets 'bn' parameters were used in the search option to access Bangla tweets only. There are over 10000 total Bangla and Romanized Bangla posts in the dataset [23].

We checked for empty rows or columns, missing annotation, proper <PN> tagging (for dataset with proper nouns replaced), proper categorization etc. The resultant dataset is now both unique and error-free in terms of the abovementioned flaws.

The entire data set was divided into Bangla and Romanized Bangla sections for convenience of future research. Scripts are available from the Data set's GitHub account to do the following:

- Converting textual data into tokens
- Saving the data as tuple ([data], [label1], [label2])
- Randomly shuffling the data
- Serializing each datasheets and splitting three sets from each and making them available for public to download and un-pickle to use them in their models.

For our experiments we applied the tokenizing, splitting, serializing scripts on the "full-text" (or unmodified texts column of the dataset with all the proper nouns, emoticons etc intact) also, hence creating additional sets of pickle files.

## V. MODEL IMPLEMENTATION

Our dataset consists of three categories –
- Positive,
- Negative, and
- Ambiguous.

Depending on the dataset used and number of categories classified, we used three types of fully connected neural networks layer, which mainly differ by the number of nodes in the output layer (Fig. 1). One and two output nodes were used for categorizing between positive and negative sentiments, and three output nodes were used when ambiguous labels were also included.

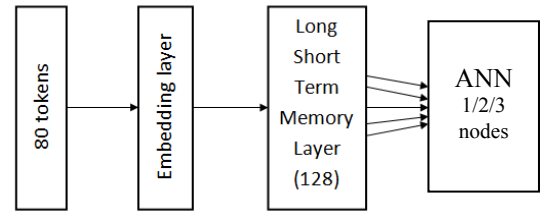

Fig. 1. Dense layer schematic

We used data for one validation set as pre-training for another validation set. More specifically, first we fit data from 1st validation in the model to pre-train for 2nd validation data – which is fit in the same model afterwards. Likewise, we fit data from 2nd validation to pre-train for 1st validation data. This sort of pre-training was to check whether it can be useful to pre-train on an independently sentiment analysis data even if the labels did not match.

## VI. EXPERIMENTS

### A. Datase Preparation

Our model is based on Recurrent Neural Networks (RNN) – more specifically we used LSTM neural network. We used Keras' model-level library since it has all the required features to help us develop our deep learning model. We used Theano as the back-end for Keras. All our models are Keras Sequential models. First layer of the Sequential model is the Embedding layer. We used Embedding layer to implement the word to vector representation for the words in our dataset. We a *max_features* parameter as the input dimension argument for embedding layer. It signifies the total number of unique tokens returned by the tokenizer, which in turn means that max_features is also the vocabulary size (*input_dim*). The second layer is Long Short Term Memory (LSTM) with an internal state of 128 dimensions. The third is a fully connected layer with different activations for classification purposes.

In our experiments, the last layer has 1, 2 and 3 nodes, depending on the classification regime attempted. When attempting to classify only positive and negative sentiment, which are represented by 1 and 0 respectively, the final fully connected layer has been configured with 1 and 2 neurons. When a single neuron is used, the loss function employed was binary-cross-entropy. When 2 neurons were used, we used categorical-cross-entropy instead. With the inclusion of ambiguous labels, the number of classes increase to 3, for which we used 3 neurons with a softmax activation in the last layer. The loss function employed in this case was categorical-cross-entropy.

Even in this simply configuration, the number of parameters of the network is quite high and it is possible to easily overfit such models. To avoid this we used Dropout [24] rates of 20% between the Embedding and LSTM layer at training time..

### B. Experiment model label tags

There are actually 36 unique experiments using the same LSTM model, depending on the dataset used, processing of texts, loss function used, processing of labels (annotations on data), and *input_dim* value for Embedding layer. However, it turns into a total of 72 experiments – one half of experiments where label 1 (1st validation) is used for pre-training, and the other half where label 2 (2nd validation) is used for pre-training. Following are the tags used in experiments and what they actually mean.

- Tags used for different types of dataset –

| Dataset Type | Tag used in experimental labels |
|---|---|
| Bangla and Romanized Bangla (total) | BRBT |
| Bangla (only) | Bangla |
| Romanized Bangla (only) | RB |

- Tags used depending on processing of texts/posts –

| Processing of texts | Tag used in experimental labels |
|---|---|
| <PN> removed and other modifications | PN |
| Full texts (no modification) | FT |

- Tags used based on loss function –

| Loss function used | Tag used in experimental labels |
|---|---|
| Binary_crossentropy | bin |
| Categorical_crossentropy | cat |

- Tags used based on Annotation data modification-

| Annotation data modification | Tag used in experimental labels |
|---|---|
| Annotation value of 'A' removed (label along with data removed) | ra |
| Annotations value of 'A' converted to 2 | ato2 |

- Tags used based on different type of max_features applied -

| Max_features type | Tag used in experimental labels |
|---|---|
| Non-fixed, ranging from 20,000 ~ 40,000 depending on the dataset type and size | 1 |
| Value fixed at 500 | 2 |

## VII. RESULTS AND DISCUSSION

Highest accuracy was attained by Bangla dataset with categorical crossentropy loss, modified text, Ambiguous removed and non-fixed max_features, with 78% of accuracy – which is 28% more than chance for two category dataset. However, this experiment on BRBT dataset with categorical loss, modified text, ambiguous converted to 2, has a low accuracy score of 55% but for a three category it scores 22% more than chance (33%). Therefore, it is clear that most of experiment sets (dataset-wise, or PN-FT tag-wise, or loss function-wise, and label category-wise) scored above chance. However, none of the experiments with fixed max_features (vocabulary size for Embedding layer) scored well compared to the non-fixed variants.

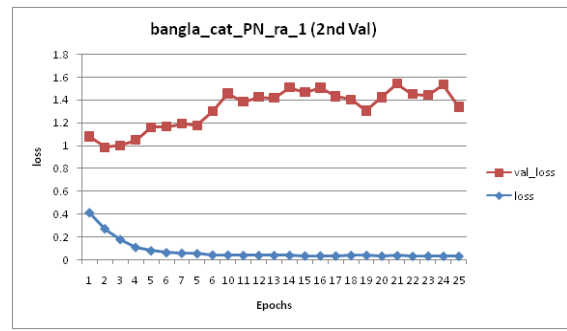

Fig. 2. loss-val_loss graph for bangla_cat_PN_ra_1 (2nd validation)

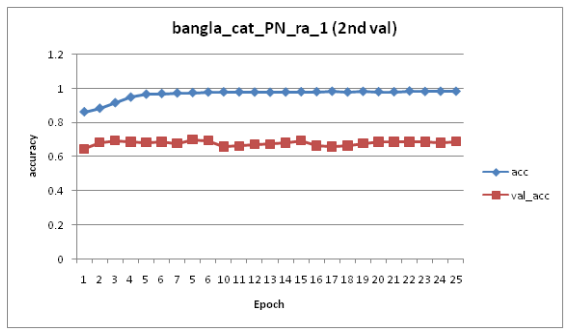

Fig. 3. acc-val_acc graph for bangla_cat_PN_ra_1 (2nd validation)

VIII. CONCLUSION

The goals of the research may be summarized as following:

1. Pre-processing the data in a way so that it is readily usable by researchers.
2. Application of deep recurrent models on a Bangla and Romanized Bangla text corpus.
3. Pre-train dataset of one label for another (vice versa) to prove its usefulness.

To meet the goals, a BRBT (Bangla and Romanized Bangla Text) dataset of total 9337 entries with 6698 entries for Bangla and 2639 for Romanized Bangla texts were pre-processed. Then dataset was split and serialized into training set, testing set and validation set.

For the experiments, LSTM which is a deep recurrent model was applied. There are total 32 different experiments based on the same model with only differences in dataset used, loss function applied, modification done (or not) on data (proper noun replaced with <PN> tags, duplication removal etc.) etc. While most of the experiments scored accuracy higher than chance in percentage, Bangla dataset with categorical crossentropy as loss function and non-fixed max_features for the embedding layer with "Ambiguous removed" scored highest with 78% in accuracy for 2 category (results compared from both pre-training set of experiments), and Bangla and Romanized Bangla dataset (modified text set) with categorical crossentropy loss, non-fixed max_features, and "Ambiguous converted to 2" scored highest with 55% in accuracy for 3 category.

The implementation of pre-training dataset of one label for another has showed that, even if the labels do not match it is useful to pre-train on an independently annotated SA data. From four experiments done from the alternate experiment set consistent result from 2nd validation data (label 2) have been observed.